\documentclass[runningheads]{llncs}

\PassOptionsToPackage{dvipsnames,table}{xcolor}
\usepackage{eccv}

\usepackage{eccvabbrv}

\usepackage{graphicx}
\usepackage{booktabs}
\usepackage{amsmath,amsfonts,amssymb}
\usepackage{algorithm}
\usepackage{algorithmic}
\usepackage{placeins}
\usepackage{array}
\usepackage{multirow}
\usepackage{textcomp}
\usepackage{makecell}
\newsavebox{\figAbox}

\usepackage{hyperref}
\usepackage{subcaption}
\usepackage{orcidlink}

\begin{document}

\title{Multi-History-Step SDE Inversion for Image Editing with Superior Regional Awareness}

\titlerunning{Multi-History-Step SDE Inversion for Region-Aware Editing}

\author{Haiyan Wei\inst{1,2} \and
Yunlong Wang\inst{1} \and
Huaibo Huang\inst{1} \and
Zhenan Sun\inst{1} \and
Kunbo Zhang\inst{1}\thanks{Corresponding author.}}

\authorrunning{H.~Wei et al.}

\institute{
New Laboratory of Pattern Recognition (NLPR), Institute of Automation, Chinese Academy of Sciences\\
\and
University of Chinese Academy of Sciences\\
\email{\{weihaiyan2026,yunlong.wang,kunbo.zhang\}@ia.ac.cn, huaibo.huang@cripac.ia.ac.cn, znsun@nlpr.ia.ac.cn}}

\maketitle

\begin{abstract}
In recent years, diffusion stochastic differential equation (SDE) inversion and inversion-free methods have become prevalent for training-free image editing, as they can achieve faithful reconstruction without tuning. However, existing approaches remain inefficient, exhibit limited plasticity, and struggle to accurately preserve unedited regions. To address these issues, we propose MIEdit, a training-free editing framework based on SDE inversion. MIEdit introduces a predictor--corrector multi-history-step scheme to achieve superior editing quality with fewer steps. We further mitigate heterogeneity and conflict between the multi-conditioned noise residuals and gradient terms during sampling, improving stability and editing plasticity under large edits. MIEdit also includes Inversion-Time Automatic Semantic Angle Masking (IASM); it leverages classifier-free guidance to automatically generate semantic angle masks during inversion and applies them throughout the sampling process for regional constraints, without extra user inputs. We additionally construct EditEval++ (30 fine-grained tasks, 1,000+ image--text--mask triplets) for comprehensive evaluation; experiments show that MIEdit outperforms state-of-the-art techniques. Project page: \url{https://whywwwzzzg.github.io/MIEdit/}.
\end{abstract}

\section{Introduction}
\label{sec:intro}

Recently, pretrained text-to-image diffusion models have driven numerous training-free image editing methods~\cite{lipman2023flowmatching,liu2022rectifiedflow,esser2024rft,blackforestlabs2024flux,zhang2024open,sun2026towards,ge2026expand}. Many image editing methods rely on deterministic ordinary differential equation (ODE) process inversion, such as Denoising Diffusion Implicit Models (DDIM) inversion and flow inversion~\cite{song2021ddim,liu2022rectifiedflow,esser2024rft,routh2024rfinversion}. These approaches map an input image back to latent variables or a noise trajectory of the diffusion process, enabling the synthesis of an edited image under a new prompt~\cite{dong2023pti,li2023stylediffusion}. However, ODE inversion typically involves implicit equations over its own variables, and explicit approximations may introduce severe reconstruction errors, particularly when conditioning is incorporated~\cite{bao2025freeinv,kulikov2025flowedit,kim2025flowalign}; consequently, many methods require tuning or multiple network evaluations within a single inversion step to enhance reconstruction consistency, which is highly time-consuming~\cite{mokady2023nulltext,chun2023exactinv,garibi2024renoise,samuel2023lightning}. Recently, approaches based on model prediction differences, including inversion-free editing and denoising diffusion probabilistic model (DDPM) inversion, have emerged, theoretically enabling error-free reconstruction, avoiding optimization and often achieving better background preservation~\cite{xu2024infedit,brack2024leditspp}.

\begin{figure}[tb]
\centering
\setlength{\abovecaptionskip}{3pt}
\setlength{\belowcaptionskip}{-7pt}
\includegraphics[width=0.50\linewidth,trim=1cm 18.35cm 1cm 1cm,clip]{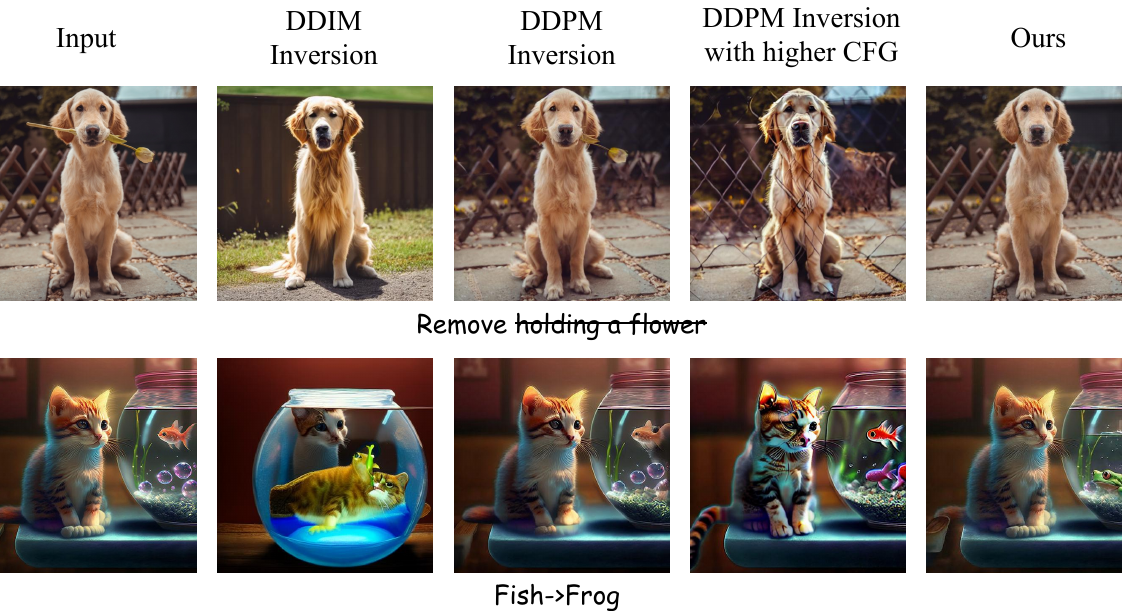}
\caption{Editing results under different methods and settings. DDIM inversion (ODE inversion) often suffers from poor consistency, while DDPM inversion (SDE inversion) may have limited plasticity: increasing CFG can degrade image quality before improving prompt adherence.}
\label{fig_overview}
\end{figure}

We generalize DDPM inversion as a form of stochastic differential equation (SDE) inversion. It leverages latents precomputed from the forward process to compute the noise terms required for reconstruction (sampling), thereby correcting the generation process~\cite{wu2023cyclediffusion,huberman2024editfriendly,brack2024leditspp}. Nevertheless, existing SDE inversion methods mainly focus on low-order SDE samplers, leaving room for improvement in both convergence speed and generation quality~\cite{huberman2024editfriendly}. In particular, the generation quality (i.e., editing plasticity) of SDE-inversion-based editing sometimes suffers from insufficient stability and may even exhibit artifacts, as shown in Fig.~\ref{fig_overview}. Moreover, both ODE- and SDE-inversion-based editing methods remain weak in preserving non-edited regions, making it difficult to precisely handle editing tasks with different modification extents~\cite{hertz2022p2p,cao2023masactrl,xu2024infedit,long2025followyourshape}. Some mask-control strategies attempt to alleviate this issue, yet they often require additional inputs or an independent mask branch~\cite{couairon2023diffedit,zhu2025kvedit}.

To address these issues, we propose \textbf{MIEdit}, an image editing method based on arbitrary multi-history-step (at least two history terms per step) diffusion SDE inversion that integrates a predictor--corrector mechanism for higher-quality image generation. MIEdit can reconstruct the input exactly (up to numerical precision) and requires no tuning. Compared to existing SDE inversion methods, our approach generates diverse samples of higher quality with fewer timesteps. Another key improvement is that we carefully modify the inference-time noise schedule, which substantially improves the generation quality of naive multi-history-step inversion in large-edit scenarios and effectively mitigates the artifact issue of SDE inversion.

To improve non-edited region control, we further propose Inversion-Time Automatic Semantic Angle Masking (IASM) without any external conditions. Specifically, we move mask generation from a byproduct of sampling to the early phase of inversion, and introduce a direction-based metric to obtain fine-grained mask control that is available throughout all generation steps. Importantly, the proposed mask does not rely on image quality during generation, nor on the gradual structural alignment between the inversion and generation stages. Our contributions are summarized as follows:
\begin{itemize}
    \item We reformulate DDPM inversion under a unified SDE inversion framework, and propose a predictor--corrector two-stage, multi-history-step SDE-based inversion-and-generation image editing method, improving both the efficiency and generation quality of current training-free image editing.
    \item We address condition-induced mismatch and interference between the noise part and gradient terms, significantly improving the plasticity of multi-history-step SDE inversion.
    \item We propose an inversion-time automatic semantic angle masking that leverages the classifier-free
guidance (CFG) process within inversion itself to produce accurate semantic masks for guiding the entire generation process. It requires no extra inputs or inference branches.
    \item We introduce a new benchmark, EditEval++, which expands EditEval's original 7 tasks~\cite{huang2024dmbie_survey} to 30 fine-grained editing tasks and contains 1,000+ image--text--mask triplets, featuring more diverse scenes and descriptions. Experiments on PIE-Bench and EditEval++ demonstrate the advantages of our method.
\end{itemize}

\section{Related Work}

\subsection{Diffusion Inversion for Image Editing}

Many training-free methods enable controllable image editing via diffusion inversion~\cite{pixartsigma2024,peebles2023dit,podell2024sdxl,esser2024rft,blackforestlabs2024flux}. The key idea is to map an image back to initial noise or intermediate states at selected timesteps so that sampling from the same latent reconstructs the input, dating back to DDIM~\cite{song2021ddim}. Methods like RF-Inv further adopt a flow-based ODE inversion strategy~\cite{routh2024rfinversion,xu2025unveil}. Faithful reconstruction, however, often requires extra optimization or multiple function evaluations per step for error correction~\cite{mokady2023nulltext,miyake2023npi,wallace2023edict,wang2024belm,bao2025freeinv}. Methods like CycleDiffusion instead adopt DDPM inversion by recovering a sequence of noise vectors such that the image can be exactly reconstructed by the DDPM sampling process~\cite{wu2023cyclediffusion,huberman2024editfriendly}. These SDE methods are closely related to recent inversion-free approaches~\cite{xu2024infedit,kulikov2025flowedit,kim2025flowalign}, where network evaluations apply differential corrections to adjust the original trajectory or image~\cite{huberman2024editfriendly}, mitigating error accumulation in iterative ODE inversion and improving reconstruction fidelity. Our method, inspired by the high-order SA-Solver~\cite{xue2023sasolver}, achieves higher efficiency and quality with exact reconstruction. SA-Solver admits a closed-form arbitrary-order SDE formulation, with DDPM and SDE-DPM-Solver++ as low-order cases~\cite{xue2023sasolver,lu2023dpmsolverpp,brack2024leditspp}.

\subsection{Precisely Localized Image Editing}
Image editing typically requires localized changes while keeping the rest of the image intact. Prior work often reuses inversion/reconstruction cues during generation (e.g., attention-feature injection~\cite{zhu2025kvedit} and blended diffusion~\cite{avrahami2023blended}), which often requires specifying regions using masks. Some studies rely on user-provided external masks, which incur substantial annotation overhead~\cite{avrahami2023blended,zhu2025kvedit,lugmayr2022repaint}. Other methods infer masks from attention maps during diffusion~\cite{hertz2022p2p,tumanyan2022pnp,cao2023masactrl,xu2024infedit,wei2025patch}. Nevertheless, this strategy typically requires an external ``blended-words'' prompt to select which token(s)' attention to use, and it often identifies the editing subject rather than the precise editing region. Moreover, existing automated mask generation methods either cannot reliably provide high-quality control throughout the full sampling trajectory, or incur additional overhead when implemented as an extra inference branch~\cite{long2025followyourshape,couairon2023diffedit,sun2025marmot}.

\section{Methods}

\subsection{Diffusion SDE and Its Inversion}

In the DDPM framework \cite{ho2020ddpm}, we are given a noise schedule $\{(\alpha_t,\sigma_t)\}_{t=0}^T$ with $\alpha_t>0$ and $\sigma_t>0$. The forward diffusion perturbs a clean sample $x_0$ into noise; its marginal form is:
\begin{equation}
x_t = \alpha_t x_0 + \sigma_t \epsilon_t, \qquad \epsilon_t \sim \mathcal{N}(0,\mathbf{I}).
\end{equation}

The corresponding reverse sampling (generation) process starts from $x_T\sim\mathcal N(0,I)$ and iterates $t=T,\dots,1$:
\begin{equation}
\label{eq:2}
x_{t-1}=\mu_t(x_t,c)+\tilde\sigma_t z_t,\quad z_t\sim\mathcal N(0,I).
\end{equation}
which yields $x_0$ at the end. Here $\mu_t(\cdot,c)$ is computed from a denoising network, and $\tilde\sigma_t$ is determined by the sampler and schedule.

In the continuous limit, the diffusion can be written as a linear SDE~\cite{song2021scoresde}:
\begin{equation}
\mathrm{d}x_t = f(t)\,x_t\,\mathrm{d}t + g(t)\,\mathrm{d}w_t,
\qquad
x_t\mid x_0 \sim \mathcal{N}\!\left(\alpha_t x_0,\sigma_t^2\mathbf{I}\right).
\end{equation}
The equivalent reverse-time generative process is:
\begin{equation}
\mathrm{d}x_t
=
\left(f(t)x_t - g(t)^2 \nabla_x \log p_t(x_t)\right)\mathrm{d}t
+
g(t)\,\mathrm{d}\bar{w}_t .
\end{equation}
We approximate the score by $s_\theta(x_t,t)$. A common parameterization is noise prediction
$\epsilon_\theta(x_t,t) = -\sigma_t s_\theta(x_t,t)$, with the corresponding data prediction:
\begin{equation}
x_\theta(x_t,t) = \frac{x_t - \sigma_t \epsilon_\theta(x_t,t)}{\alpha_t}.
\end{equation}

Based on the above, inversion aims to construct latent codes for a given image $x_0$ such that the reverse trajectory reconstructs $x_0$. For a particular SDE generation, under a given noise level and condition $c$, the initial random state and the random drives injected throughout the process (in the DDPM case, the per-step noises $\{z_t\}_{t=1}^{T}$ in Eq.~\eqref{eq:2}) uniquely determine the final generated sample $x_0$. Therefore, they can be viewed as latent codes. Correspondingly, DDPM inversion aims to, given a sample $x_0$, construct such latent variables.

\subsection{Naive SA-Inversion}

We further improve the efficiency of DDPM inversion~\cite{ho2020ddpm,huberman2024editfriendly} by optimizing the sampling strategy. 
Specifically, we rephrase DDPM inversion as SDE inversion and derive an efficient inversion scheme using a refined multi-history-step stochastic diffusion sampler together with its sampling recursion~\cite{xue2023sasolver}.

Under the SDE formulation, the latent code to be inverted corresponds to a Wiener process path $w_{[0,T]}$. In practical numerical implementation, it can be equivalently represented as the Brownian increment sequence $\{\Delta w_i\}$ on a discrete time grid. This sequence can deterministically reconstruct the input image from the noise endpoint. If we further replace the conditioning text, we obtain the corresponding image editing result.

For our method, given an input image $x_0$, we choose timesteps $\{t_i\}_{i=0}^M$ and construct an auxiliary noisy reconstruction sequence:

\begin{equation}
x^{\mathrm{inv}}_{t_i}=\alpha_{t_i}x_0+\sigma_{t_i}\tilde\epsilon,\quad \tilde\epsilon\sim\mathcal N(0,I).
\end{equation}
where $\alpha_{t_i}$ and $\sigma_{t_i}$ are respectively the signal coefficient and the noise standard deviation at time $t_i$. We define $\lambda_t=\log(\alpha_t/\sigma_t)$ and $\lambda_i \triangleq \lambda(t_i)$.

Taking the predictor in SA-Solver~\cite{xue2023sasolver} as an example (the corrector is analogous), we compute and store the per-step noise term $\eta_i^{P}$:
\begin{equation}
\eta_i^{P}
=
x_{t_{i+1}}^{\mathrm{inv}}
-
\frac{\sigma_{t_{i+1}}}{\sigma_{t_i}}
\exp\!\left(
-\int_{\lambda_i}^{\lambda_{i+1}} \tau^2(\bar{\lambda})\,\mathrm{d}\bar{\lambda}
\right)
x_{t_i}^{\mathrm{inv}}
-
\sum_{j=0}^{s-1} b_{i-j}\,x_\theta\!\left(x_{t_{i-j}}^{\mathrm{inv}},t_{i-j}\right).
\end{equation}
Here $x_\theta(\cdot,\cdot)$ is the data-prediction model; $s$ is the number of model evaluations used by the multi-history-step method (i.e., its order). In this setting, ``multi-history-step'' indicates that the current sampling step uses not only the model output from the current latent variable, but also the outputs from the preceding $s-1$ time steps for interpolation. $\tau(\cdot)$ is a variance control function; and the coefficients $b_{i-j}$ are integral weights derived from the Lagrange interpolation basis functions~\cite{xue2023sasolver}.
During generation, we use the stored per-step noise term $\eta_i^{P}$ from the inversion process to replace the random noise component $\tilde{\sigma}_i\xi$:
\begin{equation}
\label{eq:6}
x_{t_{i+1}}
=
\frac{\sigma_{t_{i+1}}}{\sigma_{t_i}}
\exp\!\left(
-\int_{\lambda_i}^{\lambda_{i+1}} \tau^2(\bar{\lambda})\,\mathrm{d}\bar{\lambda}
\right)
x_{t_i}
+
\sum_{j=0}^{s-1} b_{i-j}\,x_\theta(x_{t_{i-j}},t_{i-j})
+
\eta_i^{P}.
\end{equation}
The original noise scale is
\begin{equation}\label{eq:noise_scale}
\tilde{\sigma}_i
=
\sigma_{t_{i+1}}
\sqrt{
1-\exp\!\left(
-2\int_{\lambda_i}^{\lambda_{i+1}} \tau^2(\bar{\lambda})\,\mathrm{d}\bar{\lambda}
\right)
}.
\end{equation}

Overall pseudocode for the process is provided in the supplementary material.

\begin{figure}[tb]
\centering
\setlength{\abovecaptionskip}{3pt}
\setlength{\belowcaptionskip}{-5pt}
{\def\svgwidth{0.94\columnwidth}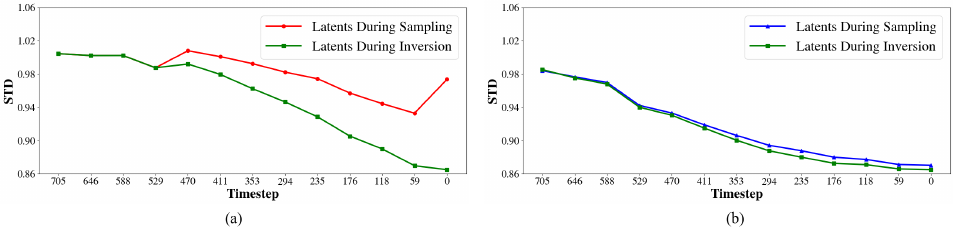}
\caption{(a) Latent standard deviation at each timestep during inversion and sampling with Naive SA-Inversion.
(b) Latent standard deviation at each timestep with our method, which prevents latent-scale explosion and keeps the trajectory within a reasonable range by closely following inversion statistics.}
\label{fig:huitucxdb}
\end{figure}

\subsection{Granting Higher Plasticity to Multi-History-Step SDE Inversion}
\label{sec:3_3}

We observe that although the Naive SA-Inversion method achieves nearly perfect reconstruction and requires only a few iterations, its editing plasticity---\ie, the ability to alter the output as instructed---is insufficient.
Specifically, when the desired edit deviates significantly from the source image, the generation result often responds weakly to the condition, and may even produce artifacts.
We examine the variance of the latent trajectory $\{x_{t_i}\}_{i=0}^{M}$ and find that it diverges from the forward noising distribution, especially at middle-to-late timesteps: the latent scale becomes biased, and the latent standard deviation is close to $1$, as shown in Fig.~\ref{fig:huitucxdb}.
The more historical steps used in SDE inversion, the more pronounced this deviation tends to be.

The key issue stems from the \emph{``stored-and-reused''} $\eta_i^{P}$.
In standard generation, the noise term follows a zero-mean Gaussian distribution with standard deviation $\tilde{\sigma}_i$ given in \eqref{eq:noise_scale}.
In contrast, during image editing, the noise term $\eta_i^{P}$ exhibits a higher variance in practice due to encompassing conditional information correlated with the source image.
While multi-step history terms improve fitting accuracy, they also introduce heterogeneity in the noise term across different latent and timestep conditions. On the other hand, when the target image differs significantly from the source image, the correlation between the noise term and the gradient term becomes weaker. These two aspects lead to superposition and conflict between the multi-step noise term and the multi-step gradient term, causing 
$x_{t_{i+1}}$ computed by Eq.~\eqref{eq:6} to have abnormally large variance.
This inflated variance then propagates to later steps and can ultimately cause the final sample to collapse. With the predictor step, we can rewrite:

\begin{equation}
\eta_i^{P}
=
x^{\mathrm{inv}}_{t_{i+1}}
-
a_i\, x^{\mathrm{inv}}_{t_i}
-
g_i ,
\end{equation}
where
\begin{equation}
a_i
=
\frac{\sigma_{t_{i+1}}}{\sigma_{t_i}}
\exp\!\left(
-\int_{\lambda_i}^{\lambda_{i+1}} \tau^2(\bar{\lambda})\, d\bar{\lambda}
\right),
\quad
g_i
=
\sum_{j=0}^{s-1}
b_{i-j}\, x_\theta\!\left(x^{\mathrm{inv}}_{t_{i-j}},\, t_{i-j}\right).
\end{equation}
Then
\begin{align}
\mathrm{Var}(\eta_i^{P})
&=
\mathrm{Var}\!\left(x^{\mathrm{inv}}_{t_{i+1}}\right)
+
a_i^2\, \mathrm{Var}\!\left(x^{\mathrm{inv}}_{t_i}\right)
+
\mathrm{Var}(g_i)
-2a_i\, \mathrm{Cov}\!\left(x^{\mathrm{inv}}_{t_{i+1}},\, x^{\mathrm{inv}}_{t_i}\right)
\nonumber\\
&\quad
-2\, \mathrm{Cov}\!\left(x^{\mathrm{inv}}_{t_{i+1}},\, g_i\right)
+2a_i\, \mathrm{Cov}\!\left(x^{\mathrm{inv}}_{t_i},\, g_i\right).
\end{align}

It can be seen that $\mathrm{Var}(\eta_i^{P})$ is mainly governed by: (1) the endpoint variances,
(2) $\mathrm{Var}(g_i)$ (whose scale is strongly tied to $\sigma_{t_{i+1}}$), and
(3) the covariance terms, among which $-2a_i\,\mathrm{Cov}(x^{\mathrm{inv}}_{t_{i+1}},x^{\mathrm{inv}}_{t_i})$ is often dominant.

Therefore, to reduce the overall variance and stabilize $\eta_i^{P}$, we lower the noise level at each timestep: this simultaneously decreases $\mathrm{Var}(x^{\mathrm{inv}}_{t_{i+1}})$ and $\mathrm{Var}(x^{\mathrm{inv}}_{t_i})$; moreover, a lower noise level also reduces $\mathrm{Var}(g_i)$ through its strong dependence on $\sigma_{t_{i+1}}$; and it increases the correlation between adjacent latents, thereby decreasing the dominant covariance contribution $-2a_i\,\mathrm{Cov}\!\left(x^{\mathrm{inv}}_{t_{i+1}},\,x^{\mathrm{inv}}_{t_i}\right)$.

\begin{figure}[tb]
\centering
{\def\svgwidth{0.94\columnwidth}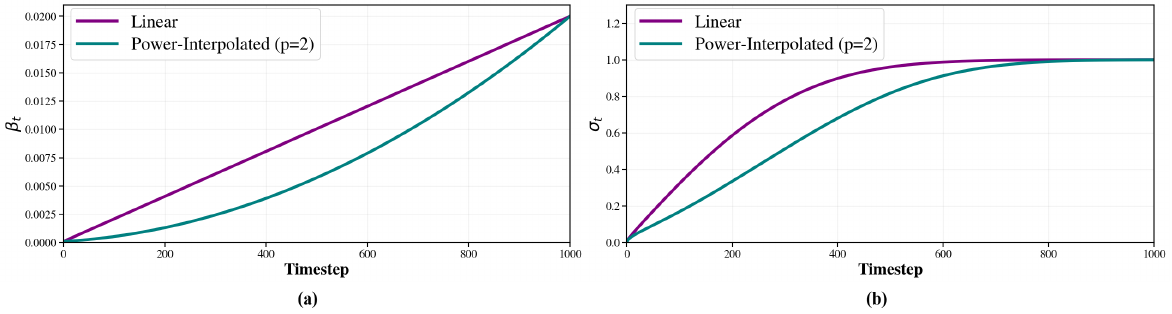}
\caption{(a) The linear $\beta$ schedule used in other inversion methods, and our power-interpolated $\beta$ schedule. (b) The corresponding $\sigma$ schedules for the two $\beta$ schedules (Eq.~\eqref{eq:sigma_schedule}), indicating the noise level at each timestep during inversion and sampling. Our power-interpolated $\beta$ schedule yields a lower noise level at the same timestep.}
\label{fig:betaschedule}
\end{figure}

Thus, we adopt a simple yet effective approach: we replace the standard linear $\beta$ schedule (used to define the forward process) with a power-interpolated schedule at inference time:
\begin{equation}
\beta_t
=
\left(
(1-u)\,\beta_{\mathrm{start}}^{1/p}
+
u\,\beta_{\mathrm{end}}^{1/p}
\right)^{p},
\qquad
p\ge 1,
\end{equation}
where $p$ is a shape-controlling parameter and $u\in[0,1]$ is the normalized time.
This schedule injects noise more smoothly at early timesteps and yields a smaller $\beta_{t_i}$ at the same timestep, as illustrated in Fig.~\ref{fig:betaschedule}.
We have:
\begin{equation}
\label{eq:sigma_schedule}
\sigma_{t_i}
=
\sqrt{1-\prod_{j=1}^{t_i}(1-\beta_j)} .
\end{equation}
This results in a smaller $\sigma_{t_i}$ at the same timestep compared to a linear schedule. It makes the sampling trajectory better match the training-time noise statistics. Ultimately, it retains perfect reconstruction (because the noise schedule used for inversion is still the same as the one used for sampling) while improving the editing plasticity of multi-history-step SDE inversion under large edits and reducing artifacts in the generated images.

\begin{figure*}[tb]
  \centering
  \includegraphics[width=\textwidth]{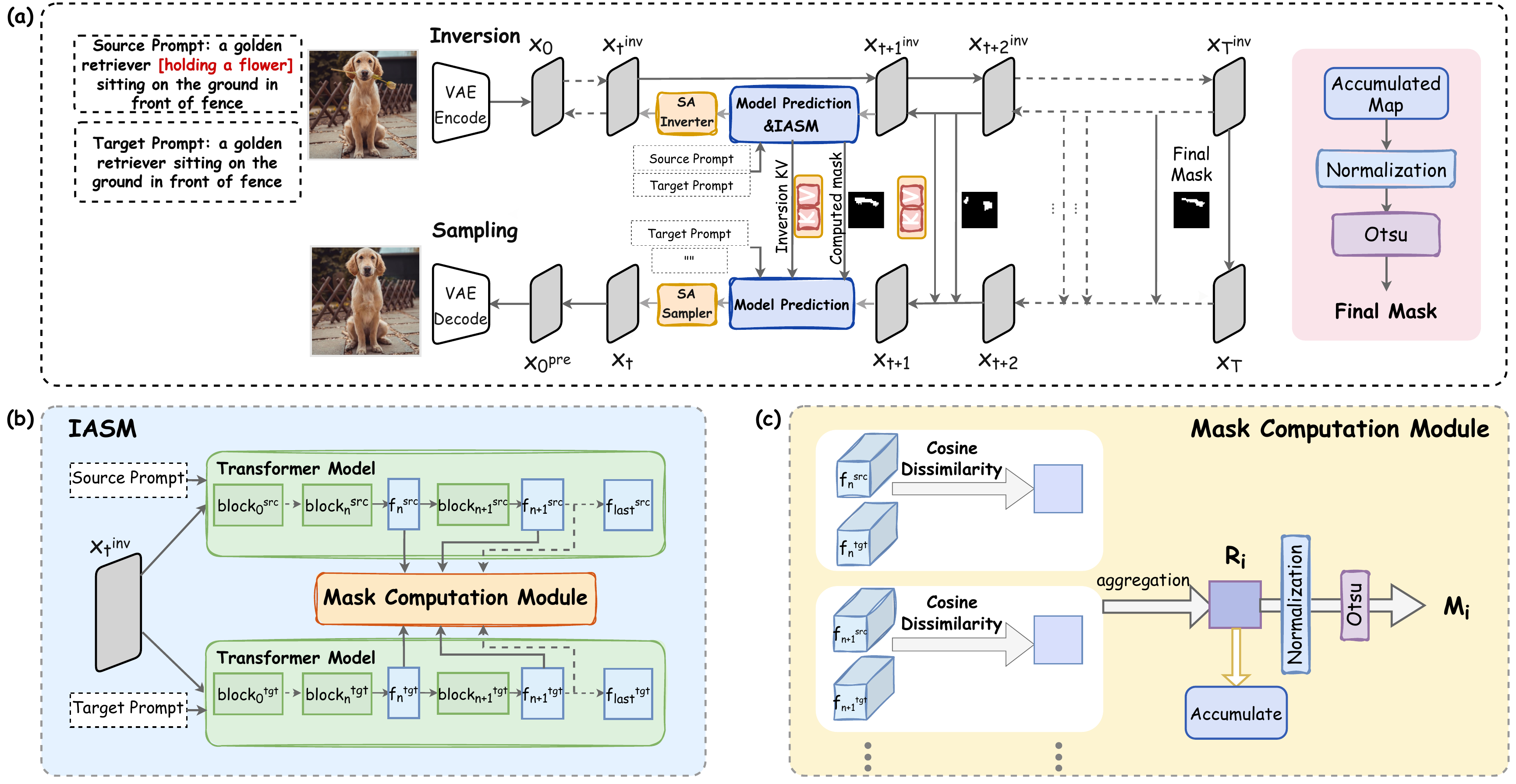}
  \caption{(a) Pipeline overview. Given a source image and prompt, we invert to obtain step-wise noise terms, masks, and key--value (KV) features, which guide generation.
(b) IASM Module. Using CFG during inversion, we set the inversion negative prompt to the target prompt for mask computation.
(c) Mask Computation Module. We compute a semantic angle mask via cosine dissimilarity between corresponding feature maps.}
  \label{fig:huitukjia}
\end{figure*}

\subsection{Inversion-Time Automatic Semantic Masking}
Most image-editing pipelines derive masks from external cues, such as user-drawn regions or blended-word tokens for attention masking~\cite{hertz2022p2p,xu2024infedit}. In addition, attention masks can be noisy, and edited regions may not align with explicit prompt tokens, making blended-word specification infeasible. Recent methods attempt to auto-generate masks without manual annotation, yet limitations persist. A common approach derives a mask from the prediction discrepancy between inversion and generation at matched timesteps~\cite{long2025followyourshape}. This method produces reliable masks only at a few mid-stage steps (early steps are too noisy), and fails to constrain early/mid generation, so small errors accumulate. Moreover, masks generated in the early and mid stages tend to be of low quality because the latent variables still contain substantial noise at these steps.

To address the need for adaptive region control, we propose \textit{Inversion-Time Automatic Semantic Masking (IASM)}. We generate masks during inversion, particularly at early and mid-stage inversion timesteps, to guide generation using inversion-derived information. Such masks originate from lower-noise latents, providing higher-confidence indications of the editing regions, and can precisely control the entire generation process, including the early generation stages. Our method requires no external conditions and remains effective even under methods with weak inversion--generation consistency; an overview is shown in Fig.~\ref{fig:huitukjia}.

\subsubsection{CFG for Masking}
For SDE inversion and inversion-free methods, classifier-free guidance (CFG)~\cite{ho2022cfg} is widely used to strengthen text conditioning by combining unconditional and conditional predictions: \begin{equation} \hat{x}_{\theta}(x_{t_i}; C, C_{\text{neg}}) = x_{\theta}(x_{t_i}, C_{\text{neg}}) + w_g \left( x_{\theta}(x_{t_i}, C) - x_{\theta}(x_{t_i}, C_{\text{neg}}) \right), \end{equation} where $w_g$ denotes the guidance scale; larger values strengthen conditional consistency. Here $x_\theta$ denotes the model predictor parameterized by $\theta$. In image editing, let $C_{\text{src}}$ and $C_{\text{tgt}}$ be the source and target prompts. Inversion and generation use guidance prompts $C^i_{\text{neg}}$ and $C^g_{\text{neg}}$, with strengths $w_i$ and $w_g$. We take a first-order SDE as an example (omitting time-dependent parameters and constants), applying the inverted latent from inversion to generation yields: \begin{equation} \begin{aligned} x_{t_{i+1}} =\; &x_{t_{i+1}}^{\mathrm{inv}} + a_i\bigl(x_{t_i}-x_{t_i}^{\mathrm{inv}}\bigr) + b_i\Bigl[ w_g\, x_\theta(x_{t_i},C_{\mathrm{tgt}}) + (w_i-1)\,x_\theta(x_{t_i}^{\mathrm{inv}},C_{\mathrm{neg}}^{i}) \\ &\qquad {}- w_i\,x_\theta(x_{t_i}^{\mathrm{inv}},C_{\mathrm{src}}) - (w_g-1)\,x_\theta(x_{t_i},C_{\mathrm{neg}}^{g}) \Bigr]. \end{aligned} \end{equation} Intuitively, CFG drives the next latent toward $C_{\text{tgt}}$ and $C_{\text{neg}}^{\,i}$, and away from $C_{\text{src}}$ and $C_{\text{neg}}^{\,g}$. However, most existing methods set $C_{\text{neg}}^i$ to empty text. This can introduce ambiguity because the unconditional (empty-text) branch is implicitly used as a positive reference during inversion, which is misaligned with the editing objective. Therefore, we set the inversion-stage guidance prompt as $C_{\text{neg}}^i = C_{\text{tgt}}$, which improves semantic consistency while maintaining background preservation; more importantly, it enables mask acquisition through CFG computation within the IASM framework. Concretely, at inversion state $x^{\text{inv}}_{t_i}$ we compute conditional predictions for both branches, $x_\theta(x^{\text{inv}}_{t_i},C_{\text{src}})$ and $x_\theta(x^{\text{inv}}_{t_i},C_{\text{tgt}})$. Small differences indicate strong source--target agreement (keep unchanged as mask background), while large differences indicate semantic conflict---regions supported by the source but not by the target---and should be edited. Since this relies on the semantic gap rather than fully specified prompts, IASM remains robust to partially descriptive prompts as long as the intended edit is expressed.

\subsubsection{Temporal Scheduling of Masks.}
We compute masks mainly on lower-noise latents in early inversion, where structures and details are clearer. Furthermore, latents at different noise intensities exhibit distinct generation focuses, and network prediction differences at corresponding noise levels indicate editing priorities for those noise scales.

We adopt temporally aligned mask scheduling in IASM. We compute single-step masks only at a few high-confidence timesteps within $[d_s, d_e)$ and apply them to the corresponding sampling steps. For the remaining (higher-noise) steps, we use an averaged mask $M_{\text{avg}}$ obtained by averaging dissimilarity maps over $[d_s, d_e)$, which reduces variance and improves stability. Formally, given single-step masks obtained at timesteps $d_s$ through $d_e-1$, the mask used at generation step $t$ is:
\begin{equation}
M_t^{\text{gen}} =
\begin{cases}
M_t, & d_s \leq t < d_e, \\
M_{\text{avg}}, & \text{otherwise},
\end{cases}
\end{equation}
This lets masks from clearer intermediate representations be used where most beneficial and stably guide early generation. 


\subsubsection{Mask Computation Method.}

Pixel- or latent-space prediction changes mostly capture appearance differences and may miss cases where pixels look similar but semantics differ.
Attention maps are more semantic but are often noisy and frequently depend on specific network designs. IASM combines the strengths by comparing internal multi-layer feature-map differences between the two network branches. On the other hand, since L2/absolute differences mainly reflect magnitude changes while semantic gaps often appear as directional changes, we use \textbf{cosine dissimilarity} to measure semantic direction consistency. At inversion state $x^{\text{inv}}_{t_i}$, we extract feature maps under source/target conditions:
\begin{equation}
f^{\text{src}}_k=F_k(x^{\text{inv}}_{t_i},C_{\text{src}}),\quad
f^{\text{tgt}}_k=F_k(x^{\text{inv}}_{t_i},C_{\text{tgt}}).
\end{equation}
where $F_k(\cdot)$ denotes the feature extractor for layer $k$, and $f_k$ represents the spatial feature map. To measure semantic conflict, we compute the cosine similarity and define the angle-response map:
\begin{equation}
S_{\text{cos}}^{(k)}(i,j)=
\frac{f_k^{\text{src}}(i,j)\cdot f_k^{\text{tgt}}(i,j)}
{\|f_k^{\text{src}}(i,j)\|_2\,\|f_k^{\text{tgt}}(i,j)\|_2},
\qquad
M_{\text{angle}}^{(k)}(i,j)=1-S_{\text{cos}}^{(k)}(i,j).
\end{equation}
Larger values indicate greater directional inconsistency and thus a higher likelihood that the position belongs to the edit region. Averaging over layers gives the dissimilarity map $R_i$ for step $i$, and applying post-processing to $R_i$ yields the single-step mask $M_i$~\cite{krahenbuhl2011densecrf}.

\section{Experiments}

\subsection{Setup}
\textbf{Baselines and Implementation.} We compare our method with representative training-free image editing approaches:
(i) feature-control methods (P2P~\cite{hertz2022p2p}, PnP~\cite{tumanyan2022pnp}, MasaCtrl~\cite{cao2023masactrl});
(ii) SDE-inversion and inversion-free methods (CycleDiffusion~\cite{wu2023cyclediffusion}, Edit-Friendly DDPM Inversion (EF)~\cite{huberman2024editfriendly}, LEDITS++~\cite{brack2024leditspp}, InfEdit~\cite{xu2024infedit}, FlowEdit~\cite{kulikov2025flowedit}, FlowAlign~\cite{kim2025flowalign}, FlowCycle~\cite{wang2025flowcycle});
and (iii) ODE-inversion methods (DDIM Inversion~\cite{song2021ddim}, Direct Inversion(DI)~\cite{ju2023directinversion}, FTEdit~\cite{xu2025unveil}, 
iRFDS~\cite{yang2024text}, FreeInv~\cite{bao2025freeinv}).
For fairness, we follow the official evaluation protocols of all baselines; for FlowEdit, we set CFG to 7.0 during editing to improve background consistency, while keeping other settings identical to its official implementation.
Our method is implemented on PixArt-$\Sigma$~\cite{pixartsigma2024}, a DiT-based model~\cite{peebles2023dit} that substantially improves over PixArt-$\alpha$~\cite{chen2024pixartalpha}.
We set CFG to 3.1/9.0 for inversion/editing, and use a shape-control parameter $p=1.9$.
To evaluate stronger backbones and ensure fair comparison with recent methods, we also implement our approach on Stable Diffusion 3 Medium and Stable Diffusion 3.5 Medium~\cite{esser2024rft}, using CFG 1.6/3.6 for inversion/editing. All experiments are conducted in PyTorch.

\textbf{Evaluation Datasets and Metrics.} We evaluate general editing performance on PIE-Bench~\cite{ju2023directinversion} and EditEval~\cite{huang2024dmbie_survey}.
PIE-Bench contains 700 natural and artificial images covering 9 dimensions; each sample provides source/target prompts, and region masks to assess background preservation and local edits.
EditEval includes 150 high-quality images across 7 task types with text prompts, but its coverage is limited and it provides no region masks for quantitative evaluation; compared to EditEval, PIE-Bench has slightly lower resolution ($512\times512$) and visual quality, and a higher proportion of artistic images.

\begin{figure}[tb]
\centering
{\def\svgwidth{0.85\columnwidth}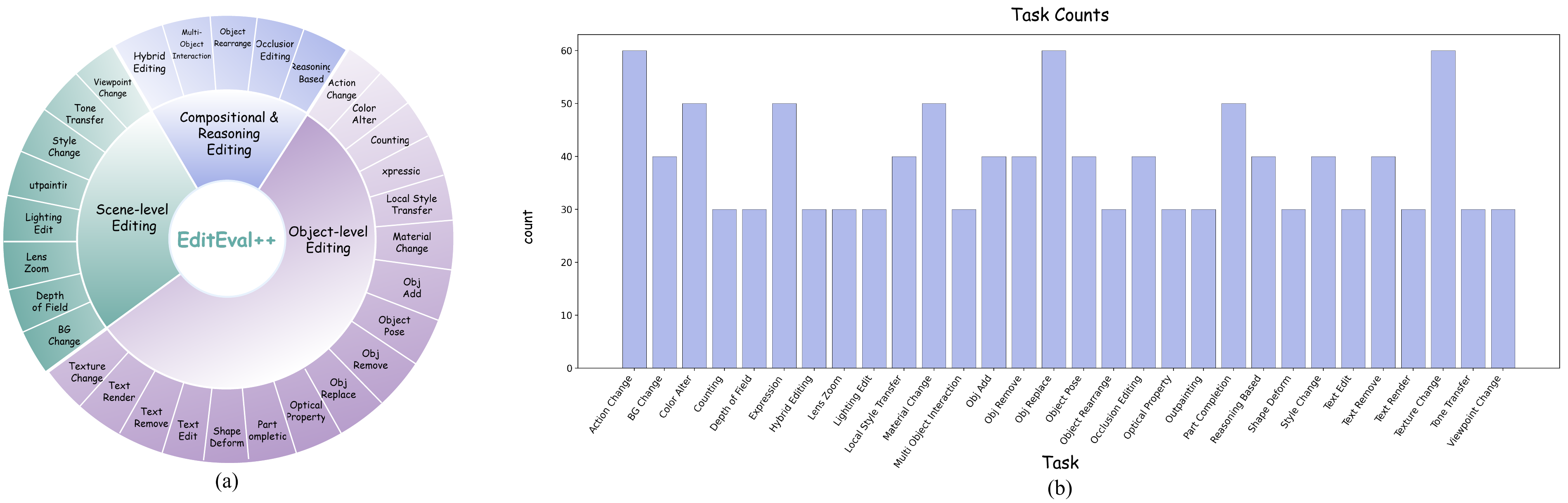}
\caption{(a) We propose EditEval++, which covers 30 tasks across three major categories: Object-level Editing, Scene-level Editing, and Compositional \& Reasoning Editing. (b) The number of (image, prompt, mask) triplets for each task.}
\label{fig:benchmark}
\end{figure}
\begin{figure*}[tb]
  \centering
  {\def\svgwidth{0.85\textwidth}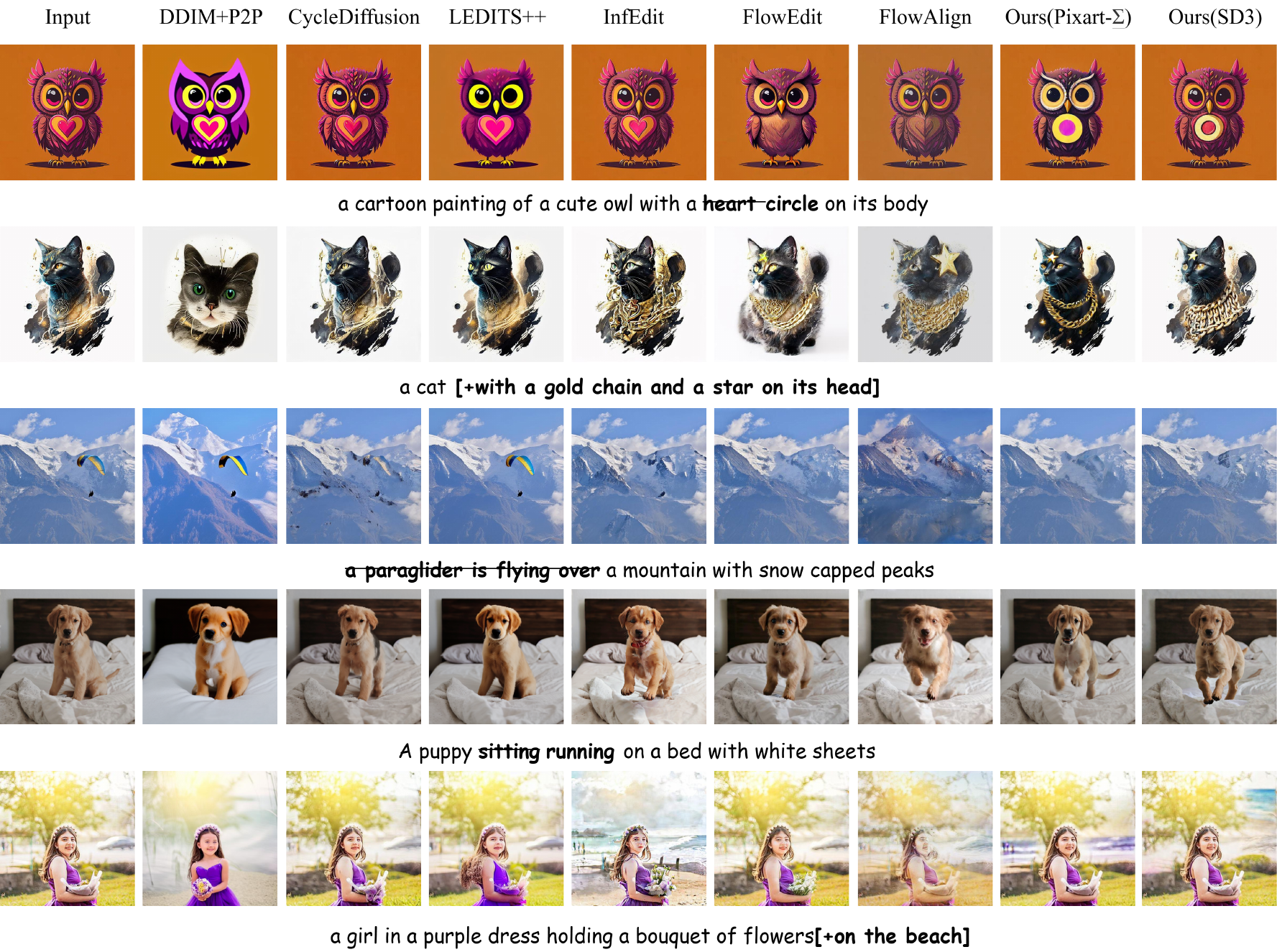}
  \caption{Qualitative results of different methods.}
  \label{fig:qualitative}
\end{figure*}

Therefore, we propose \textbf{EditEval++}, a more challenging large-scale image editing benchmark that expands EditEval's original 7 tasks to 30 fine-grained tasks with varying difficulty levels---positioning it among the most comprehensive description-based image editing benchmarks currently available, as shown in Fig.~\ref{fig:benchmark}. Other benchmarks remain limited: TedBench~\cite{kawar2023imagic} covers only 4 tasks with 100 samples, and EditVal~\cite{basu2023editval} spans 12 tasks but offers fewer samples and is instruction-based rather than description-based. EditEval++ includes classic tasks such as shape changes, object additions, and challenging operations that push the limits of existing methods, including lighting edits and depth-of-field modifications. We expand the sample size from EditEval's 150 to 1160 samples and provide corresponding masks for each sample (for evaluation purposes only). The detailed construction pipeline is provided in the supplementary material.

For metrics, following PIE-Bench, we measure non-edited region preservation using structure distance~\cite{ju2023directinversion}, PSNR, LPIPS~\cite{zhang2018unreasonable}, MSE, and SSIM~\cite{wang2004image}, and measure edit success using CLIP similarity~\cite{radford2021learning} between the edited region and the target text prompt. Some metrics in the tables are rescaled. For EditEval++, we assess alignment with human preferences via EditScore~\cite{luo2026editscore}. We also report the number of function evaluations (NFE) for computational efficiency. For fairness, we define one NFE as one forward pass on a single sample; for example, methods that perform both reconstruction and generation incur two NFEs per step, since the actual computational cost is doubled.

\begin{table*}[tb]
\caption{Quantitative comparison on PIE-Bench. (best: bold; second: underlined)}
\label{tab:piebench}
\centering
\scriptsize
\setlength{\tabcolsep}{2.15pt}
\renewcommand{\arraystretch}{1.2}
\begin{tabular}{l l c c c c c c c}
\toprule
\multirow{2}{*}{Method} & \multirow{2}{*}{Model} &
\multicolumn{1}{c}{Structure} &
\multicolumn{4}{c}{Background Preservation} &
\multirow{2}{*}{\makecell{CLIP\\Similarity$\uparrow$}} &
\multirow{2}{*}{NFE$\downarrow$} \\
\cmidrule(lr){3-3}\cmidrule(lr){4-7}
& & Distance$\downarrow$ & PSNR$\uparrow$ & LPIPS$\downarrow$ & MSE$\downarrow$ & SSIM$\uparrow$ &  &  \\
\midrule
DDIM+P2P & SD1.4 & 69.43 & 17.87 & 208.80 & 219.88 & 71.14 & 22.44 & 150 \\
CycleDiffusion & SD1.4 & \underline{11.15} & 26.93 & 62.43 & 35.41 & 84.29 & 21.68 & 118 \\
DI+MasaCtrl & SD1.4 & 23.58 & 22.68 & 87.41 & 80.63 & 81.51 & 21.41 & 200 \\
DI+PnP & SD1.5 & 24.29 & 22.46 & 106.06 & 80.40 & 79.68 & 22.62 & 150 \\
EF+P2P & SD1.4 & 18.05 & 24.55 & 91.88 & 94.58 & 81.57 & 21.03 & 126 \\
InfEdit & SD1.4 & 14.22 & 27.52 & 47.98 & 
34.17 & 85.05 & 22.03 & 64 \\
LEDITS++ & SD1.5 & 15.76 & 24.54 & 71.25 & 53.05 & 83.08 & 20.92 & 60 \\
iRFDS & SD3 & 62.72 & 19.61 & 186.39 & 179.76 & 74.59 & 21.67 & 2816 \\
FTEdit & SD3.5 & 30.12 & 21.94 & 118.04 & 88.09 & 83.35 & \underline{22.65} & 150 \\
FlowEdit & SD3 & 12.29 & 26.56 & 55.30 & 34.23 & 89.45 & 22.50 & 66 \\
FreeInv+P2P & SD1.5 & 17.13 & 26.03 & 67.90 & 41.70 & 83.00 & 22.33 & 150 \\
FlowAlign & SD3 & 28.30 & 25.50 & 53.28 & 43.78 & 87.92 & 22.00 & 66 \\
FlowCycle & SD3 & 13.30 & 26.83 & 63.24 & 33.56 & 88.63 & 22.46 & 6633 \\
\midrule
Ours & PixArt-$\Sigma$ & \textbf{11.10} & 28.01 & 48.13 & 36.18 & 87.11 & 22.26 & \underline{48} \\
Ours & SD3 & 12.07 & \textbf{29.44} & \textbf{40.03} & \textbf{23.49} & \textbf{91.17} & 22.41 & \textbf{38} \\
Ours & SD3.5 & 13.21 & \underline{29.35} & \underline{42.36} & \underline{25.67} & \underline{90.86} & \textbf{22.73} & \textbf{38} \\
\bottomrule
\end{tabular}

\end{table*}
\subsection{Comparison with Previous Editing Methods}
We qualitatively evaluate our method with baselines on image editing tasks, with results shown in Fig.~\ref{fig:qualitative}. Our method successfully executes edits according to prompts with high fidelity while better preserving non-target content, validating its comprehensive capability for image editing. Quantitative results on PIE-Bench are summarized in Table~\ref{tab:piebench}. Our method achieves excellent performance across most metrics and outperforms all baselines in background preservation and instruction fidelity, with instruction fidelity measured by CLIP scores. In particular, our method attains substantially better background preservation than competing approaches. Notably, there exists a fundamental trade-off between preserving source content and achieving strong editing capability. Furthermore, our method demonstrates significantly higher efficiency compared to other methods. We evaluate SDE inversion and inversion-free methods on EditEval++, with results shown in Table~\ref{tab:editevalpp} and detailed task-by-task performance shown in the supplementary material. Our method maintains superior performance---achieving higher CLIP scores while better preserving background content---further validating its robustness across diverse tasks. As reported in Table~\ref{tab:editevalpp}, our method (SD3) achieves the highest EditScore of 4.30, demonstrating its superior perceptual quality and consistency with human judgment. Moreover, among the methods built on SD3, our approach is the fastest and requires only 1.8 seconds per image, as shown in Table~\ref{tab:runtime}.


\begin{table}[tb]
\caption{Quantitative comparison on EditEval++ (best in bold).}
\label{tab:editevalpp}
\centering
\scriptsize
\setlength{\tabcolsep}{2.35pt}
\renewcommand{\arraystretch}{1.2}
\begin{tabular}{lccccccc}
\hline
Method &
\makecell{Structure\\Distance}$\downarrow$ &
PSNR$\uparrow$ &
LPIPS$\downarrow$ &
MSE$\downarrow$ &
SSIM$\uparrow$ &
\makecell{CLIP\\Similarity}$\uparrow$ &
EditScore$\uparrow$ \\
\hline
CycleDiffusion &
11.15 &
28.06 &
67.13 &
33.24 &
85.98 &
22.55 &
3.66 \\
LEDITS++ &
16.06 &
25.41 &
71.19 &
50.12 &
83.96 &
22.22 &
3.09 \\
InfEdit (LCM) &
33.88 &
20.48 &
131.37 &
243.20 &
76.13 &
22.79 &
4.05 \\
FlowEdit &
11.21 &
27.31 &
58.40 &
32.64 &
89.53 &
22.62 &
3.84 \\
FlowAlign &
23.50 &
26.36 &
71.58 &
42.53 &
87.91 &
22.91 &
4.02 \\
Ours (PixArt-$\Sigma$) &
\textbf{10.75} &
28.21 &
52.21 &
37.51 &
87.75 &
22.85 &
3.98 \\
Ours (SD3) &
11.58 &
\textbf{29.58} &
\textbf{50.42} &
\textbf{25.75} &
\textbf{90.52} &
\textbf{23.30} &
\textbf{4.30} \\
\hline
\end{tabular}
\end{table}

\begin{figure}[tb]
\centering
{\def\svgwidth{0.9\columnwidth}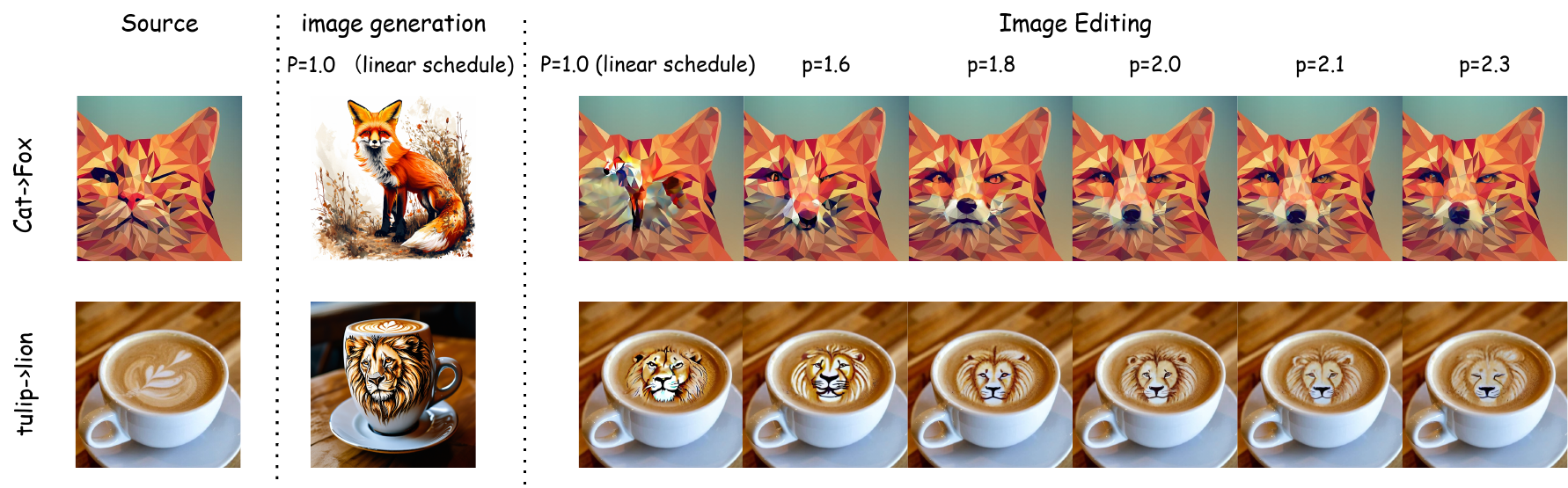}
\caption{The impact of the shape-controlling parameter $p$ on the results.}
\label{fig:ablation_p}
\end{figure}

\subsection{Ablation Study and Analysis}
We conduct ablation studies on key components of our algorithm on PIE-Bench. First, we study the role of historical terms in the stochastic linear multistep sampling and inversion scheme with integrated predictor--corrector updates.
With a fixed number of sampling steps in our PixArt-$\Sigma$ implementation, Table~\ref{tab:pc_history_terms} shows that using 2 historical terms for prediction (a third-order predictor) and 3 for correction (a fourth-order corrector) achieves the best editing performance. Next, we analyze our proposed $\beta$-schedule design by adjusting the shape-controlling parameter $p$. linear $\beta$ schedule can achieve high-quality text-to-image generation; however, for large-scale image editing, it tends to introduce artifacts. In contrast, our power-interpolated $\beta$ schedule yields stable and high-quality results. Finally, we evaluate IASM for mask localization by comparing the Attention Mask from InfEdit~\cite{xu2024infedit} and the Sampling-generated Mask from Follow-Your-Shape~\cite{long2025followyourshape} within our SD3.5-based SA-Inversion pipeline. Due to the absence of corresponding blended words in many editing prompts and the stronger structural guarantees provided by our algorithm compared to FollowYourShape, the results in Table~\ref{tab:mask_ablation} and Fig.~\ref{fig:ablation_last} demonstrate our superior capability in localizing editing regions. The last two rows of Table~\ref{tab:mask_ablation} further validate our temporal mask scheduling and the advantage of semantic angle masks over magnitude-based masks.



\begin{table}[t]
\centering
\scriptsize
\setlength{\tabcolsep}{1.6pt}
\caption{Ablations on history terms and mask design.}
\label{tab:ablation_two_tables}
\hspace*{-0.015\linewidth}\begin{subtable}[t]{0.485\linewidth}
\centering
\caption{}
\label{tab:pc_history_terms}
\renewcommand{\arraystretch}{1.49}
\begin{tabular}{lccccc}
\hline
& \multicolumn{5}{c}{$\left(\begin{array}{cc}
\text{Predictor} & \text{Corrector} \\
\text{history terms}, & \text{history terms}
\end{array}\right)$} \\
\cline{2-6}
Metric & (1, 1) & (2, 2) & (3, 2) & (2, 3) & (3, 3) \\
\hline
PSNR$\uparrow$ & 27.35 & 27.80 & 27.95 & 28.01 & \textbf{28.08} \\
\makecell[l]{CLIP$\uparrow$} & 21.73 & 22.01 & 22.06 & \textbf{22.26} & 22.17 \\
\hline
\end{tabular}
\end{subtable}\hspace{0.005\linewidth}
\begin{subtable}[t]{0.485\linewidth}
\centering
\caption{}
\label{tab:mask_ablation}
\renewcommand{\arraystretch}{1.25}
\begin{tabular}{lccc}
\hline
Method & PSNR$\uparrow$ & LPIPS$\downarrow$ & CLIP$\uparrow$ \\
\hline
\begin{tabular}{@{}l@{}}Attention Mask\\(from InfEdit)\end{tabular}  & 28.86 & 48.06 & 22.47 \\
\begin{tabular}{@{}l@{}}Sampling-generated Mask\\(from Follow-Your-Shape)\end{tabular} & 26.82 & 60.44 & 22.39 \\
\textbf{IASM (Ours)} & \textbf{29.35} & \textbf{42.36} & \textbf{22.73} \\
Full-Step IASM & 29.23 & 43.20 & 22.68 \\
Magnitude-Based Mask & 28.46 & 44.63 & 22.63 \\
\hline
\end{tabular}
\end{subtable}

\end{table}
\begin{figure*}[tb]
\centering

\begin{minipage}[t]{0.48\textwidth}
\vspace{0pt} 
\centering
\scriptsize
\setlength{\tabcolsep}{2.0pt}
\renewcommand{\arraystretch}{1.35}
\captionof{table}{Average inference time per image on NVIDIA RTX 5090. Comparison among Stable Diffusion 3 based methods.}
\label{tab:runtime}
\begin{tabular}{lcccc}
\hline
Method & FlowEdit & FlowAlign & FlowCycle & Ours \\
\hline
\makecell[l]{Time\\(s/img)} & 2.6 & 2.4 & 270.0 & \textbf{1.8} \\
\hline
\end{tabular}
\end{minipage}\hfill
\begin{minipage}[t]{0.48\textwidth}
\vspace{0pt} 
\centering
{\def\svgwidth{0.94\linewidth}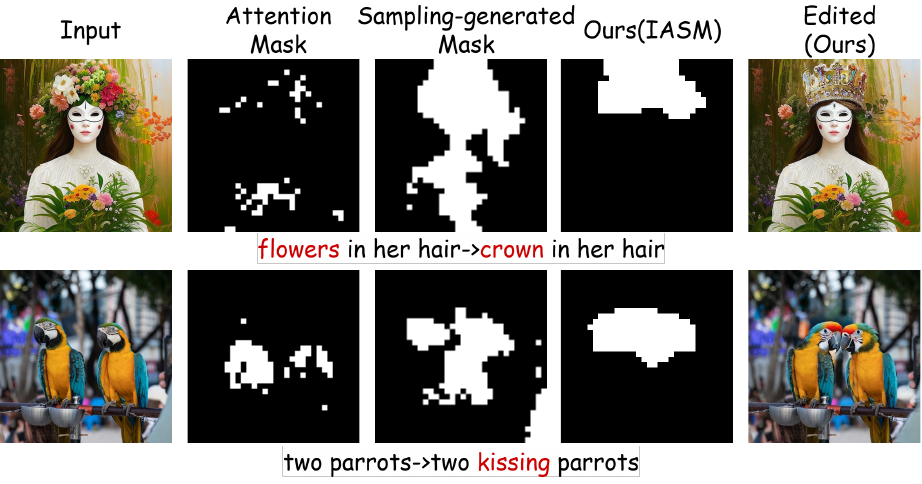}
\captionof{figure}{Comparison of three types of localization strategies.}
\label{fig:ablation_last}
\end{minipage}

\end{figure*}

\FloatBarrier
\section{Conclusion}
We present MIEdit, a training-free image editing framework based on high-order SDE inversion. A predictor--corrector multi-history scheme enables exact reconstruction with fewer steps, while an optimized inference noise schedule improves stability and quality for large edits. IASM automatically derives semantic angle masks during inversion from CFG feature discrepancies to constrain regions throughout sampling without extra inputs. Extensive experiments and the new EditEval++ benchmark demonstrate consistent gains, especially on challenging structural edits, and suggest promising extensions to video and 3D generation.

\section*{Acknowledgements}
This work was supported by the National Natural Science Foundation of China (Grant Nos. 62476272 and 62576342), and by the Tianjin Key Research and Development Program CAS-Cooperation Project (Grant No. 24YFYSHZ00290).

\bibliographystyle{splncs04}
\bibliography{refs}
\end{document}